\documentclass[11pt, a4paper]{lumia}

\usepackage[sort&compress]{natbib}
\usepackage{xspace}

\theoremstyle{plain}
\newtheorem*{proposition*}{Proposition}

\theoremstyle{definition}

\theoremstyle{definition}

\def\eqref#1{equation~\ref{#1}}
\usepackage{graphicx}           % graphics support
\usepackage{tikz}               % TikZ graphics
\usepackage[edges]{forest}      % forest trees
\usepackage{url}                % simple URL typesetting
\usepackage{xurl}               % extended URL breaking

\usepackage{array}              % extended array and tabular
\usepackage{longtable}          % multi-page tables
\usepackage{multirow}           % multirow cells
\usepackage{makecell}           % enhanced cell formatting
\usepackage{ragged2e}           % ragged text alignment

\usepackage{mathtools}          % additional math tools
\usepackage{nicefrac}           % compact symbols for 1/2, etc.

\usepackage{algorithm}          % algorithm environment
\usepackage{algorithmicx}       % algorithmic pseudocode
\usepackage{algpseudocode}      % pseudocode formatting
\usepackage{listings}           % code listings

\usepackage{subcaption}         % subfigures and subcaptions
\usepackage{wrapfig}            % text wrapping around figures
\usepackage[export]{adjustbox}  % adjustable boxes

\usepackage[commandnameprefix=always]{changes}            % track changes
\usepackage{xspace}             % intelligent spacing
\usepackage[normalem]{ulem}     % underlining without affecting emphasis
\usepackage{CJKutf8}            % CJK character support

\usepackage[tikz]{bclogo}       % logo boxes
\usepackage[framemethod=tikz]{mdframed} % framed text

\usepackage{lipsum}             % lorem ipsum text
\usepackage{tocloft}            % table of contents formatting
\usepackage{afterpage}          % page break control
\usepackage{bbding}             % additional symbols
\usepackage{epigraph}           % epigraphs
\usepackage{minitoc}            % mini table of contents
\usepackage{multicol}           % multiple columns
\usepackage{textgreek}          % Greek text

\newcolumntype{P}[1]{>{\RaggedRight\arraybackslash}p{#1}}

\definecolor{uclablue}{RGB}{39, 116, 174}
\definecolor{bigaired}{RGB}{156, 0, 0}
\definecolor{myblue}{HTML}{598BE7}
\definecolor{mildblue}{RGB}{31,119,180}
\definecolor{sectionblue}{RGB}{70, 130, 180}
\definecolor{methodblue}{RGB}{0, 150, 136}
\definecolor{bgblue}{RGB}{245,243,253}
\definecolor{ttblue}{RGB}{91,194,224}
\definecolor{mygreen}{rgb}{0.64, 0.56, 0.88}
\definecolor{myyellow}{rgb}{0.68, 0.6, 0.1}
\definecolor{fancygreen}{rgb}{0.33, 0.68, 0.20}
\definecolor{salmon}{rgb}{0.94, 0.52, 0.49}
\definecolor{tablegreen}{rgb}{0.82, 0.94, 0.75}
\definecolor{tableblue}{rgb}{0.81, 0.90, 0.94}
\definecolor{tablered}{rgb}{0.97, 0.85, 0.85}
\definecolor{tableorange}{rgb}{0.96, 0.85, 0.81}
\definecolor{myorange}{rgb}{1.0, 0.49, 0.0}
\definecolor{tlgreen}{rgb}{0.33, 0.68, 0.20}
\definecolor{darkgreen}{RGB}{0,100,0}
\definecolor{darkred}{RGB}{200, 0, 0}
\definecolor{customyellow}{HTML}{FFFACD}
\definecolor{refinegreen}{RGB}{0, 128, 75}
\definecolor{scoregreen}{RGB}{34, 139, 34}
\definecolor{hidden-blue}{RGB}{194,232,247}
\definecolor{hidden-black}{RGB}{20,68,106}
\definecolor{yes}{HTML}{C6EFCE}
\definecolor{no}{HTML}{FFC7CE}
\definecolor{partial}{HTML}{FFEB9C}
\definecolor{external}{HTML}{D9E1F2}
\definecolor{hdr}{HTML}{F2F2F2}
\definecolor{GRPOrow}{gray}{0.96}
\definecolor{FlowRLrow}{RGB}{225,236,255}
\definecolor{FlowBlue}{RGB}{80,120,210}
\definecolor{GRPOGray}{gray}{0.35}

\hypersetup{
    colorlinks=true, 
    citecolor=uclablue, 
    linkcolor=bigaired,
    urlcolor=darkblue
}

\setlist[itemize]{leftmargin=20pt, noitemsep, topsep=0pt}

\NewDocumentCommand{\kaiyan}{mO{}}{\textcolor{purple}{\textsuperscript{\textit{kaiyan}}\textsf{\textbf{\small[#1]}}}}
\NewDocumentCommand{\yuxin}{mO{}}{\textcolor{cyan}{\textsuperscript{\textit{yuxin}}\textsf{\textbf{\small[#1]}}}}
\NewDocumentCommand{\bx}{mO{}}{\textcolor{green}{\textsuperscript{\textit{bx}}\textsf{\textbf{\small[#1]}}}}
\NewDocumentCommand{\at}{mO{}}{\textcolor{red}{\textsuperscript{\textit{AT}}\textsf{\textbf{\small[#1]}}}}
\NewDocumentCommand{\re}{mO{}}{\textcolor{blue}{\textsuperscript{\textit{RE}}\textsf{\textbf{\small[#1]}}}}
\NewDocumentCommand{\ybsun}{mO{}}{\textcolor{magenta}{\textsuperscript{\textit{youbang}}\textsf{\textbf{\small[#1]}}}}
\NewDocumentCommand{\runze}{mO{}}{\textcolor{orange}{\textsuperscript{\textit{runze}}\textsf{\textbf{\small[#1]}}}}
\NewDocumentCommand{\add}{mO{}}{\textcolor{darkgreen}{\textsuperscript{\textit{Maybe Consider Discuss}}\textsf{\textbf{[#1]}}}}

\newcommand{\cmark}{\textcolor{darkgreen}{\boldmath$\checkmark$}}
\newcommand{\xmark}{\textcolor{darkred}{\boldmath$\times$}}

\newenvironment{itemize*}%
 {\leftmargini=10pt\begin{itemize}%
  \setlength{\itemsep}{0pt}%
  \setlength{\parskip}{0pt}%
  }%
 {\end{itemize}}

\newenvironment{enumerate*}%
 {\begin{enumerate}%
  \setlength{\itemsep}{0pt}%
  \setlength{\parskip}{0pt}}%
 {\end{enumerate}}

\newcommand{\cellstatus}[1]{%
  \begingroup
  \StrTrim{#1}[\statusval]%
  \IfStrEq{\statusval}{Yes}{\cellcolor{yes}\cmark}{}%
  \IfStrEq{\statusval}{No}{\cellcolor{no}\xmark}{}%
  \IfBeginWith{\statusval}{Yes (}{\cellcolor{yes}\cmark~\textit{\statusval\unskip}}{}%
  \IfStrEq{\statusval}{Partial}{\cellcolor{partial}\textbf{Partial}}{}%
  \IfStrEq{\statusval}{External}{\cellcolor{external}\textbf{External}}{}%
  \endgroup
}

\newtcolorbox{myboxi}[1][]{
  breakable,
  title=#1,
  colback=red!5,
  colbacktitle=red!5,
  coltitle=black,
  fonttitle=\bfseries,
  bottomrule=0pt,
  toprule=0pt,
  leftrule=2pt,
  rightrule=2pt,
  titlerule=0pt,
  arc=0pt,
  outer arc=0pt,
  colframe=red,
}

\newtcolorbox{myboxnote}[1][]{
  breakable,
  title=#1,
  colback=orange!0,
  colbacktitle=orange!0,
  coltitle=black,
  fonttitle=\bfseries,
  bottomrule=0pt,
  toprule=0pt,
  leftrule=2pt,
  rightrule=2pt,
  titlerule=0pt,
  arc=0pt,
  outer arc=0pt,
  colframe=orange,
}

\newtcolorbox{myboxii}[1][]{
  breakable,
  freelance,
  title=#1,
  colback=white,
  colbacktitle=white,
  coltitle=black,
  fonttitle=\bfseries,
  bottomrule=0pt,
  boxrule=0pt,
  colframe=white,
  overlay unbroken and first={
  \draw[red!75!black,line width=3pt]
    ([xshift=5pt]frame.north west) -- 
    (frame.north west) -- 
    (frame.south west);
  \draw[red!75!black,line width=3pt]
    ([xshift=-5pt]frame.north east) -- 
    (frame.north east) -- 
    (frame.south east);
  },
  overlay unbroken app={
  \draw[red!75!black,line width=3pt,line cap=rect]
    (frame.south west) -- 
    ([xshift=5pt]frame.south west);
  \draw[red!75!black,line width=3pt,line cap=rect]
    (frame.south east) -- 
    ([xshift=-5pt]frame.south east);
  },
  overlay middle and last={
  \draw[red!75!black,line width=3pt]
    (frame.north west) -- 
    (frame.south west);
  \draw[red!75!black,line width=3pt]
    (frame.north east) -- 
    (frame.south east);
  },
  overlay last app={
  \draw[red!75!black,line width=3pt,line cap=rect]
    (frame.south west) --
    ([xshift=5pt]frame.south west);
  \draw[red!75!black,line width=3pt,line cap=rect]
    (frame.south east) --
    ([xshift=-5pt]frame.south east);
  },
}

\tcbset{
  takeawaysbox/.style={
    title=Takeaways,
    colback=lightblue!80,
    colframe=black,
    fonttitle=\bfseries\small,
    coltitle=white,
    colbacktitle=black,
    enhanced,
    attach boxed title to top left={xshift=2.5mm,yshift=-2.5mm},
    boxed title style={rounded corners, size=small, colframe=black, colback=black},
    width=\linewidth,
    arc=3.5mm
  }
}

\mdfdefinestyle{mystyle}{%
  rightline=true,
  innerleftmargin=10,
  innerrightmargin=10,
  outerlinewidth=3pt,
  topline=false,
  rightline=true,
  bottomline=false,
  skipabove=\topsep,
  skipbelow=\topsep
}

\tikzset{%
    every node/.style={font=\tiny},
    parent/.style =          {align=center,text width=2cm,rounded corners=3pt, line width=0.3mm, fill=gray!10,draw=gray!80},
    child/.style =           {align=center,text width=2.0cm,rounded corners=3pt, fill=blue!10,draw=blue!80,line width=0.3mm},
    grandchild/.style =      {align=center,text width=2cm,rounded corners=3pt},
    greatgrandchild/.style = {align=center,text width=1.5cm,rounded corners=3pt},
    greatgrandchild2/.style = {align=center,text width=1.5cm,rounded corners=3pt},    
    referenceblock/.style =  {align=center,text width=1.5cm,rounded corners=2pt},
    pretrain/.style =           {align=center,text width=2.0cm,rounded corners=3pt, fill=blue!10,draw=blue!80,line width=0.3mm},   
    pretrain_work/.style =           {align=center, text width=8.5cm,rounded corners=3pt, fill=blue!10,draw=blue!0,line width=0.3mm},  
    template/.style =           {align=center,text width=2.0cm,rounded corners=3pt, fill=red!10,draw=red!80,line width=0.3mm},   
    template_work/.style =           {align=center,text width=8.5cm,rounded corners=3pt, fill=red!10,draw=red!0,line width=0.3mm},    
    answer/.style =           {align=center,text width=2.0cm,rounded corners=3pt, fill= cyan!10,draw= cyan!80,line width=0.3mm},   
    answer_work/.style =           {align=center,text width=8.5cm,rounded corners=3pt, fill= cyan!10,draw= cyan!0,line width=0.3mm},      
    multiple/.style =           {align=center,text width=2.0cm,rounded corners=3pt, fill= orange!10,draw= orange!80,line width=0.3mm},   
    multiple_work/.style =           {align=center,text width=8.5cm,rounded corners=3pt, fill= orange!10,draw= orange!0,line width=0.3mm},        
    tuning/.style =           {align=center,text width=2.0cm,rounded corners=3pt, fill= magenta!10,draw= magenta!80,line width=0.3mm},   
    tuning_work/.style =           {align=center,text width=8.5cm,rounded corners=3pt, fill= magenta!10,draw= magenta!0,line width=0.3mm},          
}

\newcommand{\lstbg}[3][0pt]{{\fboxsep#1\colorbox{#2}{\strut #3}}}

\lstdefinelanguage{diff}{
  basicstyle=\ttfamily\small,
  morecomment=[f][\lstbg{red!20}]-,
  morecomment=[f][\lstbg{green!20}]+,
}

\lstdefinelanguage{diffpython}{
  language=diff,
  morekeywords={def, if, else, for, while, return, import, from, as, class, with, try, except, finally, raise, lambda, and, or, not, in, is, None, True, False},
  morecomment=[l]{\#},
  morestring=[b]",
  morestring=[b]',
}

\setheadertext{LUMIA Lab}

\correspondingemail{\emailicon lin.zhouhan@gmail.com \quad \emailicon liming\_lu@sjtu.edu.cn \quad $^\ddagger$ Corresponding Author. }

\renewcommand{\today}{2026-2-01}

\title{One Size Does Not Fit All: Token-Wise Adaptive Compression for KV Cache}
\setheadertitle{One Size Does Not Fit All: Token-Wise Adaptive Compression for KV Cache}

\author{%
  Liming Lu$^{1,3,4}$, Kaixi Qiu$^{1,3,4}$, Jiayu Zhou$^{1,5}$, Jushi Kai$^{1,4}$, Haoyan Zhang$^{2,4}$, Huanyu Wang$^{1}$, Jingwen Leng$^{4}$, Ziwei He$^{2,\ddagger}$,Zhouhan Lin$^{1,2,6,\ddagger}$\\
  $^1$ LUMIA Lab, School of Artificial Intelligence, Shanghai Jiao Tong University\\
  $^2$ Shanghai Innovation Institute\\
  $^3$ John Class, Zhiyuan College, Shanghai Jiao Tong University \\
  $^4$ School of Computer Science, Shanghai Jiao Tong University \\
  $^5$ School of Mathematical Sciences, Shanghai Jiao Tong University \\
  $^6$ Shanghai Artificial Intelligence Laboratory
  
}

\begin{document}

% ====================
% ABSTRACT
% ====================
\begin{abstract}
Despite the remarkable progress of Large Language Models (LLMs), the escalating memory footprint of the Key-Value (KV) cache remains a critical bottleneck for efficient inference. While dimensionality reduction offers a promising compression avenue, existing approaches typically either necessitate prohibitively expensive pre-training from scratch or suffer from severe performance deterioration under high compression regimes. In this work, we propose DynaKV, a novel post-training framework for low-rank KV cache compression. To the best of our knowledge, DynaKV is the first method to dynamically allocate compression rates to individual tokens according to their semantic meaning, which allows it to achieve better fidelity at aggressive compression ratios. Extensive experiments demonstrate that our method consistently outperforms existing state-of-the-art compression techniques, achieving significant memory reduction while maintaining competitive generation quality. Furthermore, our approach is orthogonal to sequence-level pruning methods. When integrated with SnapKV, DynaKV retains only 6\% of the KV cache while maintaining 94\% of the baseline performance on the LongBench benchmark.\footnote{We will release our code for reproducibility later.}
\end{abstract}

% Generate title with LUMIA style formatting
\maketitle

% ====================
% YOUR PAPER CONTENT GOES HERE
% ====================

%\newpage
\section{Introduction}
% \footnote{\bai{Try proofread these paragraphs with ChatGPT/Kimi. }}
While Large Language Models (LLMs) \citep{llm} have revolutionized AI, the escalating memory overhead of the Key-Value (KV) cache has emerged as a critical bottleneck as model sizes and context lengths grow. Specifically, the memory footprint of cached states grows linearly with sequence length, rapidly exhausting device memory. This excessive overhead imposes a major challenge on scalability, effectively preventing the deployment of larger-scale models and severely limiting the generation of extended contexts.

To mitigate this issue, extensive research has been conducted to compress the KV cache. One prominent direction focuses on employing compact low-rank representations for the KV pairs. Approaches such as ASVD \citep{Asvd}, Palu \citep{palu} and CommonKV \citep{commonkv} obtain compact KV representations by decomposing the projection weight matrices into low-rank forms. Moreover, architectural innovations such as Multi-head Latent Attention (MLA) \citep{deepseek} embed information into lower-dimensional subspaces to achieve inherent compression.

Existing methods face a dilemma between performance and training overhead. Architectural solutions like MLA require training the model from scratch on massive datasets, making them prohibitively expensive to apply to existing pre-trained LLMs. Although methods like MHA2MLA \citep{mha2mla} allow converting existing MHA models to MLA, they still necessitate Billions of training tokens to make it work. Alternatively, training-free and Post-training compression methods, such as Palu \citep{palu} and MatryoshkaKV \citep{matryoshkakv}, Despite their advantage of low adaptation costs, they inherently trade off performance for efficiency. Consequently, the primary challenge for these training-free and post-training approaches lies in sustaining performance at high compression ratios.

% 请将这段代码放在两个自然段落之间
\begin{wrapfigure}{r}{0.6\columnwidth} % {r}=靠右, {0.6\columnwidth}=预留总宽(保持您原图的大小)
    \centering
    \vspace{-10pt} % 【可选】向上调整位置，消除顶部空白
    
    % 注意：这里必须用 \linewidth，让图片填满上面设定的 0.6\columnwidth
    \includegraphics[width=\linewidth]{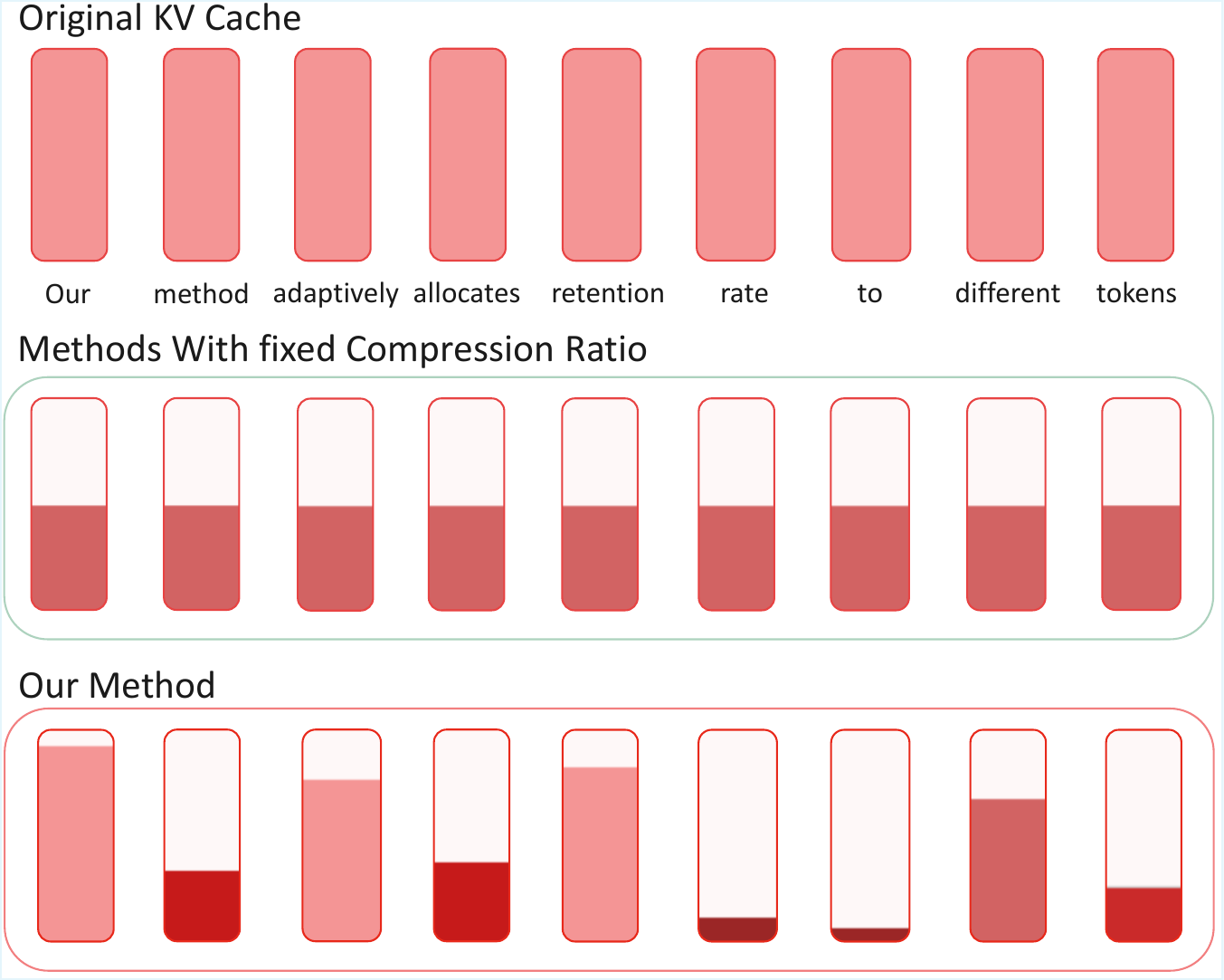}
    
    \vspace{-5pt} % 【可选】让 Caption 离图片近一点
    \caption{Illustration of the compression strategies. The diagram compares fixed-ratio compression with our proposed dynamic approach. While traditional methods apply a uniform compression rate to all tokens, DynaKV allocates variable storage budgets, assigning different retain rate of KV to different tokens based on their importance.}
    \label{fig:overview}
    \vspace{-5pt} % 【可选】消除底部空白
\end{wrapfigure}
A critical observation motivating our work is that natural language exhibits highly non-uniform information density. In language model predictions, the majority of tokens are generated with high confidence (low entropy), indicating substantial computational redundancy while making these predictions. In addition, works like CoM \citep{CoM} have shown that different tokens need a variable amount of computation according to their importance. Despite this intrinsic variability, existing low-rank compression paradigms typically enforce a rigid, uniform compression policy across the entire sequence. This ``one-size-fits-all" strategy is inherently suboptimal: it allocates equal memory budgets to critical and trivial tokens alike, inevitably wasting resources on the latter or degrading the fidelity of the former. Theoretically, an adaptive strategy that dynamically aligns memory allocation with local semantic importance holds the promise of achieving significantly higher compression rate without compromising the model's core representational fidelity.

To materialize this potential, we propose \textbf{DynaKV}, which, to the best of our knowledge, is the first post-training framework to enable \textbf{token-adaptive} compression. Unlike prior works, our method empowers the model to \textbf{automatically allocate compression rates across different tokens}. By effectively leveraging this capability, DynaKV successfully bridges the gaps in current research, offering three distinct advantages:
\begin{itemize}
    \item \textbf{Seamless Adaptation to Pre-trained Models:} As a post-training method, DynaKV can be directly applied to existing LLM architectures without structural modifications.
    \item \textbf{Minimal Training Overhead:} Our approach requires only lightweight fine-tuning to calibrate the dynamic gating mechanism (e.g., only 128M tokens for an 8B model), avoiding the prohibitive costs of training from scratch.
    \item \textbf{High Compression with Robust Performance:} By intelligently preserving critical information, DynaKV maintains competitive generation quality even at aggressive compression ratios where uniform methods typically fail.
\end{itemize}

\section{Related Work}

\paragraph{Training-free Compression Methods}
This category of approaches aims to compress KV pairs without updating model parameters, typically by exploiting the inherent low-rank properties of weight matrices or activations. Eigen Attention \citep{eigen_attn} observes that the standard attention mechanism can be approximated by maintaining a truncated basis of the key-value subspace, effectively projecting KV pairs into a lower-dimensional manifold during inference. Similarly, PALU \citep{palu} decomposes the linear projection matrices into low-rank components via Singular Value Decomposition (SVD), reconstructing the full cache on-the-fly to reduce memory complexity to the rank size. Addressing inter-layer redundancy, CommonKV \citep{commonkv} identifies that attention features often converge across layers. It employs SVD to extract a shared basis for adjacent layers and stores only the layer-specific differences, significantly reducing redundancy with a training-free parameter sharing mechanism.

\paragraph{Post-training Compression Methods}
Post-training methods introduce lightweight optimization or calibration to adapt the model for higher compression rates. ASVD \citep{Asvd} improves upon standard decomposition by introducing activation-aware scaling, which weighs the singular values by the magnitude of input activations to preserve the accuracy of outlier features during compression. MatryoshkaKV \citep{matryoshkakv} enforces a nested structure within the representation, ensuring that the most critical information is concentrated in the initial dimensions. This allows for elastic truncation, where the cache size can be dynamically adjusted by slicing the embedding vector with minimal performance degradation. Additionally, ReCalKV \citep{recalkv} critiques prior methods for neglecting the distinct roles of Keys and Values. It introduces a tailored post-training strategy that applies Head-wise Similarity–aware Reordering (HSR) to Keys for improved grouped SVD, while employing Offline Value Calibration (OVC) to refine Value projection matrices using calibration data to ensure accurate contextual representation.

\paragraph{Architecture-Intrinsic Methods}
These methods fundamentally alter the attention architecture, necessitating pre-training from scratch or computationally expensive retraining. GQA (Grouped Query Attention) \citep{gqa} mitigates the memory bottleneck by grouping multiple query heads to share a single key-value head, offering a balanced trade-off between the quality of Multi-Head Attention (MHA) \citep{attention} and the efficiency of Multi-Query Attention (MQA) \citep{mqa}. MLA (Multi-Head Latent Attention) \citep{deepseek}, introduced in DeepSeek-V2, natively projects KV pairs into a low-dimensional latent space and decouples positional embeddings, maximizing compression capabilities while maintaining model capacity. While MHA2MLA \citep{mha2mla} attempts to bridge the gap by converting existing MHA models into the MLA format, it relies on matrix decomposition followed by extensive training (e.g., billions of tokens) to recover performance, making the computational cost comparable to partial pre-training.
\begin{figure*}[t] % [t] 表示强制尝试放在页首
  \centering
  % width=0.95\linewidth 保证图片占满版心宽度但留一点点白边，视觉更舒适
  % 请将 'your_figure_filename.pdf' 替换为您实际的文件名
  \includegraphics[width=1\linewidth]{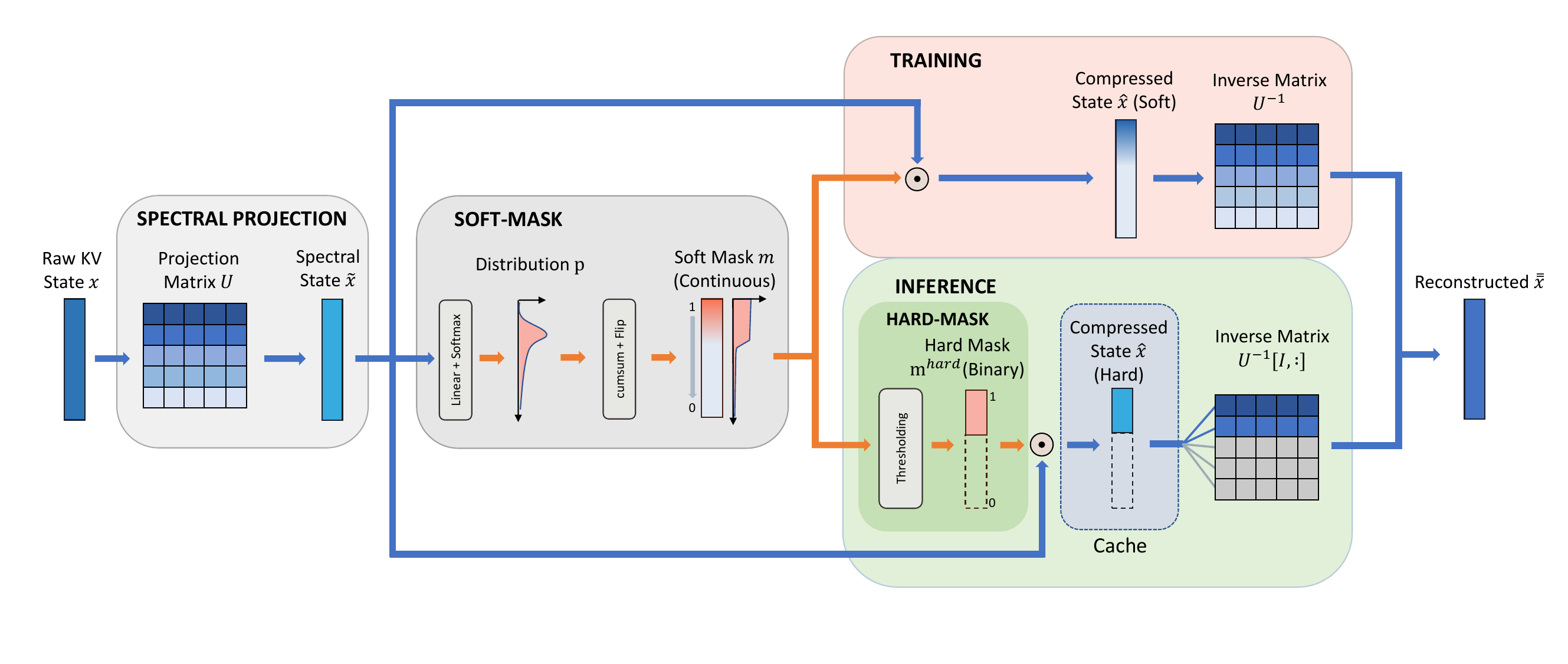}
  
  \vspace{-1em} % 如果图片和标题距离太远，可以用这个负值拉近一点
  
  \caption{\textbf{Overview of the DynaKV framework.} 
  Unlike static methods that use a uniform compression rate, DynaKV employs a token-adaptive masking mechanism to dynamically select and retain critical KV dimensions. 
  This ensures that semantically significant context is preserved while redundancy is minimized, maintaining high performance across both short and long-context tasks.}
  \label{fig:main_overview}
  
  \vspace{-1em} % 减少标题下方的空白，节省版面
\end{figure*}

\section{Dynamic Compressing}
\label{sec:method}

In this section, we present \textbf{DynaKV}, a novel framework for token-adaptive KV cache compression. Our approach is founded on the observation that the information distribution within the KV representations is highly non-uniform and varies significantly across different tokens. The proposed method consists of three key components: (1) a basis transformation via Principal Component Analysis (PCA) to decouple feature correlations, (2) a learnable, differentiable gating mechanism that dynamically determines the optimal retention rank for each token, and (3) a compression-incentivizing training objective.

\subsection{KV Projection to Spectral Space}
\label{subsec:pca}

We aim to project $\mathbf{x}$ into a compact spectral space by transforming its coordinate axes, such that the dimensions are strictly ordered by their importance. In this spectral space, the trailing dimensions exhibit minimal significance, allowing us to preferentially prune them during the subsequent compression phase with negligible information loss.

Let $\mathbf{x} \in \mathbb{R}^{d_{kv}}$ denote the pre-RoPE Key and Value states for the $t$-th token. Formally, we introduce a learnable projection matrix $\mathbf{U} \in \mathbb{R}^{d_{kv} \times d_{kv}}$ to transform the original state:
\begin{equation}
    \tilde{\mathbf{x}} = \mathbf{x} \mathbf{U}.
\end{equation}
In this transformed space, $\mathbf{U}$ is optimized to concentrate the semantic energy of the vector into the leading dimensions, providing a natural inductive bias that ensures the tail dimensions are safe to prune. To establish this property from the outset, we initialize $\mathbf{U}$ as the eigen matrix of PCA derived from offline calibration\footnote{Specifically, we construct a calibration dataset by collecting sufficient large number of consecutive tokens from standard text sources such as Wikipedia. We perform PCA on the covariance matrix of these states and use the resulting eigenvectors (sorted by descending eigenvalues) to initialize $\mathbf{U}$.}, and subsequently allow it to be updated during training to adaptively align with the compression objective.

\subsection{Differentiable Token-Adaptive Compression}
\label{subsec:masking}

To achieve token-adaptive compression, we introduce a learnable gating mechanism that dynamically determines the information retention level for each token. In this section, we first detail how this mechanism is applied during inference to realize physical storage reduction and state reconstruction. Subsequently, we describe the differentiable training process that enables the model to learn these optimal truncation points end-to-end.

\paragraph{Hard Masking at Inference Time}
During the inference phase, our goal is to minimize KV cache usage by physically discarding redundant spectral components. A key property of our transformed space is that dimensions are strictly ordered by importance, allowing us to prune dimensions starting from the tail end. Accordingly, for each token in each layer, we utilize a binary truncation mask $\mathbf{m}^{hard}\in \mathbb{R}^{d_{kv}}$, which is a vector starting with a contiguous block of $\mathbf{1}$s followed by $\mathbf{0}$s. The specific generation mechanism for $\mathbf{m}^{hard}$ is detailed in the subsequent section. Formally, the truncation is applied to the spectral state via the Hadamard product:
\begin{equation}
    \hat{\mathbf{x}} = \tilde{\mathbf{x}} \odot \mathbf{m}^{hard}.
\end{equation}
In terms of physical storage, we denote the set of retained indices as $\mathcal{I} = \{j \mid m^{hard}_j = 1\}$. Only the components of $\hat{\mathbf{x}}$ indexed by $\mathcal{I}$ are explicitly persisted in the KV cache, thereby realizing the actual memory reduction.

When reconstructing the stored KV, since different tokens retain varying numbers of components, this is implemented as a dynamic low-rank reconstruction:
\begin{equation}
    \bar{\bar{\mathbf{x}}} = \hat{\mathbf{x}}[\mathcal{I}] \mathbf{U}^{-1}[\mathcal{I}, :].
\end{equation}
Here, $\bar{\bar{\mathbf{x}}}$ represents the reconstructed Key or Value vector. Following reconstruction, these states are reshaped into the multi-head format, and Rotary Positional Embeddings (RoPE) are applied to the Key vectors.

\paragraph{Retain Rate}
To quantify and control the global compression level, we define the \textit{Retain Rate} $\mathcal{R}$ as the average proportion of retained dimensions. Let $m^{hard}_i$ denote the $i$-th entry of vector $\mathbf{m}^{hard}$. The \textit{Retain Rate} is defined as:
\begin{equation}
    \mathcal{R} = \frac{1}{d_{kv}} \sum m^{hard}_i.
\end{equation}
It is important to note that $\mathcal{R}$ here is specific to a single token within a particular layer. To calculate the model's overall retention rate, we average $R$ across the entire token sequence, all layers, and all batches.

\paragraph{Differentiable Masking at Training Time}
To generate the binary truncation mask $\mathbf{m}^{hard}$, we optimize a continuous soft-mask $\mathbf{m}$ that serves as a learnable proxy for the discrete truncation operation.

Drawing inspiration from the latent structure modeling in PRPN \citep{prpn} and the ordered representation in ON-LSTM \citep{ordered}, we implement a differentiable gating mechanism using a smooth relaxation. For a transformed spectral state $\tilde{\mathbf{x}}$, we first employ a lightweight linear layer to project the features into a probability distribution over possible truncation indices. The trainable parameters $\mathbf{W}$ and $\mathbf{b}$ produce logits which are normalized via Softmax:
\begin{equation}
    \mathbf{p} = \text{Softmax}(\tilde{\mathbf{x}} \mathbf{W} + \mathbf{b}).
\end{equation}
Here, the vector $\mathbf{p}$ represents the probability distribution of the optimal cutoff point across all dimensions. Crucially, we utilize separate sets of projection parameters for Keys and Values to accommodate their distinct redundancy patterns.

To construct a mask that starts at $\mathbf{1}$ (retained) and transitions to $\mathbf{0}$ (discarded), we leverage the cumulative sum operation. Intuitively, if the network predicts a high cutoff probability at index $k$, the cumulative sum will transition from 0 to 1 at that index. We apply a reversal operation to invert this sequence:
\begin{equation}
    \mathbf{m} = \text{Flip}\left( \text{cumsum}(\mathbf{p}) \right).
\end{equation}
The resulting mask $\mathbf{m}$ is approximately 1 for indices $i < k$ and 0 for $i > k$, providing a smooth gradient flow for backpropagation. This generated soft-mask then serves as a \textbf{differentiable gate}, filtering the spectral features via element-wise multiplication:
\begin{equation}
    \hat{\mathbf{x}} = \tilde{\mathbf{x}} \odot \mathbf{m}.
\end{equation}
During training, we retain the continuous values of $\mathbf{m}$ to maintain gradient flow, allowing the model to learn the optimal trade-off between memory usage and performance.

During inference, however, we discretize this continuous gate to achieve physical compression. Specifically, we obtain the binary mask $\mathbf{m}^{hard}$ by applying a pre-defined threshold $\tau$ (e.g., $\tau=0.1$) to the learned soft mask:
\begin{equation}
    m^{hard}_i = \mathbb{I}(m_i > \tau),
\end{equation}
where $\mathbb{I}(\cdot)$ is the indicator function. This binary mask $\mathbf{m}^{hard}$ determines the actual truncation point, ensuring that only the informative dimensions are preserved in the KV cache while the rest are discarded.

\subsection{Training Objective}
\label{subsec:loss}

To balance the trade-off between model performance and compression efficiency, we propose a composite objective that incorporates both the standard language modeling loss and a regularization term based on the \textit{Retain Rate} $\mathcal{R}$ (as defined in Section \ref{subsec:masking}).

Let $\mathcal{L}_{CE}$ denote the standard cross-entropy loss over the target tokens. The final loss function $\mathcal{L}$ is formulated as:
\begin{equation}
    \mathcal{L} = \mathcal{L}_{CE} + \alpha \cdot \mathcal{R}^2,
\end{equation}
where $\alpha$ is a tunable hyperparameter governing the magnitude of the compression penalty. Crucially, to ensure valid gradient flow during optimization, the $\mathcal{R}$ in this objective is the averaged overall retain rate computed using the continuous differentiable mask $\mathbf{m}$.

We perform continue pre-training using this composite objective. By varying $\alpha$, we can explicitly control the equilibrium between the KV cache budget and generation quality, obtaining models with varying degrees of compression.
\section{Experiments}
\label{sec:experiments}
\begin{table}[t]
\centering
\caption{Zero-shot performance on short-context benchmarks. ``Avg'' denotes the average score across these five tasks. \textbf{Bold} indicates the best performance under comparable compression rates (excluding Full Cache).}
\label{tab:short_context}

% 使用 resizebox 适配单栏宽度
\resizebox{0.7\columnwidth}{!}{%
    \setlength{\tabcolsep}{4pt} % 稍微收紧列间距
    \renewcommand{\arraystretch}{0.8}

    \begin{tabular}{lc ccccc c}
    \toprule
    \textbf{Method} & \textbf{Rate} & \textbf{ARC-C} & \textbf{ARC-E} & \textbf{PIQA} & \textbf{Wino} & \textbf{Hella} & \textbf{Avg} \\
    \midrule
    \multicolumn{8}{c}{\textit{Llama-3-8B}} \\
    \midrule
    Full Cache & 100\% & 45.76 & 70.02 & 80.85 & 71.51 & 74.68 & 68.56 \\
    \midrule
    Palu & 50\% & 39.32 & 63.84 & 77.58 & \textbf{67.80} & 69.00 & 63.51 \\
    MatryoshkaKV & 50\% & 38.64 & 60.49 & 78.02 & 63.14 & 67.36 & 61.53 \\
    \textbf{DynaKV} & 47\% & \textbf{44.75} & \textbf{67.37} & \textbf{80.36} & 67.64 & \textbf{71.70} & \textbf{66.36} \\
    \midrule
    Palu & 30\% & 29.15 & 27.51 & 71.60 & 60.62 & 57.19 & 49.21 \\
    MatryoshkaKV & 30\% & 32.88 & 49.74 & 73.67 & 55.80 & 56.80 & 53.78 \\
    \textbf{DynaKV} & 29\% & \textbf{42.71} & \textbf{59.96} & \textbf{79.82} & \textbf{66.54} & \textbf{70.33} & \textbf{63.87} \\
    \midrule
    Palu & 20\% & 29.49 & 28.57 & 65.72 & 54.70 & 46.47 & 44.99 \\
    MatryoshkaKV & 20\% & 27.46 & 45.15 & 68.50 & 53.83 & 45.31 & 48.05 \\
    \textbf{DynaKV} & 20\% & \textbf{38.98} & \textbf{59.44} & \textbf{78.56} & \textbf{65.51} & \textbf{67.89} & \textbf{62.08} \\
    \midrule
    \multicolumn{8}{c}{\textit{Qwen3-8B-Base}} \\
    \midrule
    Full Cache & 100\% & 44.41 & 71.60 & 79.76 & 70.40 & 72.28 & 67.69 \\
    \midrule
    Palu & 50\% & 32.88 & 49.74 & 71.38 & 62.43 & 60.74 & 55.43 \\
    \textbf{DynaKV} & 48\% & \textbf{42.37} & \textbf{70.19} & \textbf{78.78} & \textbf{64.96} & \textbf{68.18} & \textbf{64.90} \\
    \midrule
    Palu & 30\% & 27.12 & 37.74 & 61.97 & 53.75 & 40.04 & 44.12 \\
    \textbf{DynaKV} & 30\% & \textbf{40.00} & \textbf{50.79} & \textbf{76.71} & \textbf{63.14} & \textbf{67.08} & \textbf{59.54} \\
    \midrule
    Palu & 20\% & 26.44 & 31.57 & 56.91 & 50.99 & 32.39 & 39.66 \\
    \textbf{DynaKV} & 18\% & \textbf{39.32} & \textbf{59.96} & \textbf{75.41} & \textbf{58.72} & \textbf{62.44} & \textbf{59.17} \\
    \bottomrule
    \end{tabular}%
}
\end{table}
\subsection{Experimental Settings}
\paragraph{Models and Benchmarks}
We evaluate DynaKV on LLaMA-3-8B \citep{llama} and Qwen3-8B-Base \citep{qwen3}. For long-context understanding, we employ LongBench \citep{longbench} and RULER \citep{ruler}. To assess general capabilities, we test on standard short-context benchmarks including ARC-C, ARC-E \citep{arc}, PIQA \citep{piqa}, Winogrande \citep{winogrande}, and HellaSwag \citep{hellaswag}. Additionally, we measure perplexity (PPL) on C4 \citep{c4} and PG-19 \citep{pg19}.

\paragraph{Baselines}
We compare our method with Palu \citep{palu}, a low-rank compression approach for KV cache. Specifically, we adopt the J-LRD variant for evaluation. We also compare against MatryoshkaKV \citep{matryoshkakv}, which employs a nested embedding structure for static truncation, serving as a fixed-rank baseline.

\paragraph{Implementation Details}
Our model is continued pre-trained on the RedPajama-V2-sample dataset \citep{redpajama} for 128 million tokens. By varifying $\alpha$ in the training loss, we obtain models with different compression rates. The average retain ratio of our models is measured on the WikiText-2 dataset \citep{wiki}. All experiments are performed on NVIDIA H200 GPUs. MatryoshkaKV is excluded from the Qwen3-8B-Base model due to compatibility issues. The extreme variance in Qwen3's specific dimensions causes the inverse Cayley transform applied in MatryoshkaKV to produce NaN values, preventing the successful construction of the compressed basis.

\subsection{Main Results}
\begin{wrapfigure}{r}{0.55\columnwidth} % {r=靠右}{占用栏宽的 55%}
    \centering
    \vspace{-5pt} % 【可选】向上微调，消除顶部多余空白
    
    % 注意：这里改成 \linewidth，表示填满上面设定的 0.55\columnwidth
    \includegraphics[width=\linewidth]{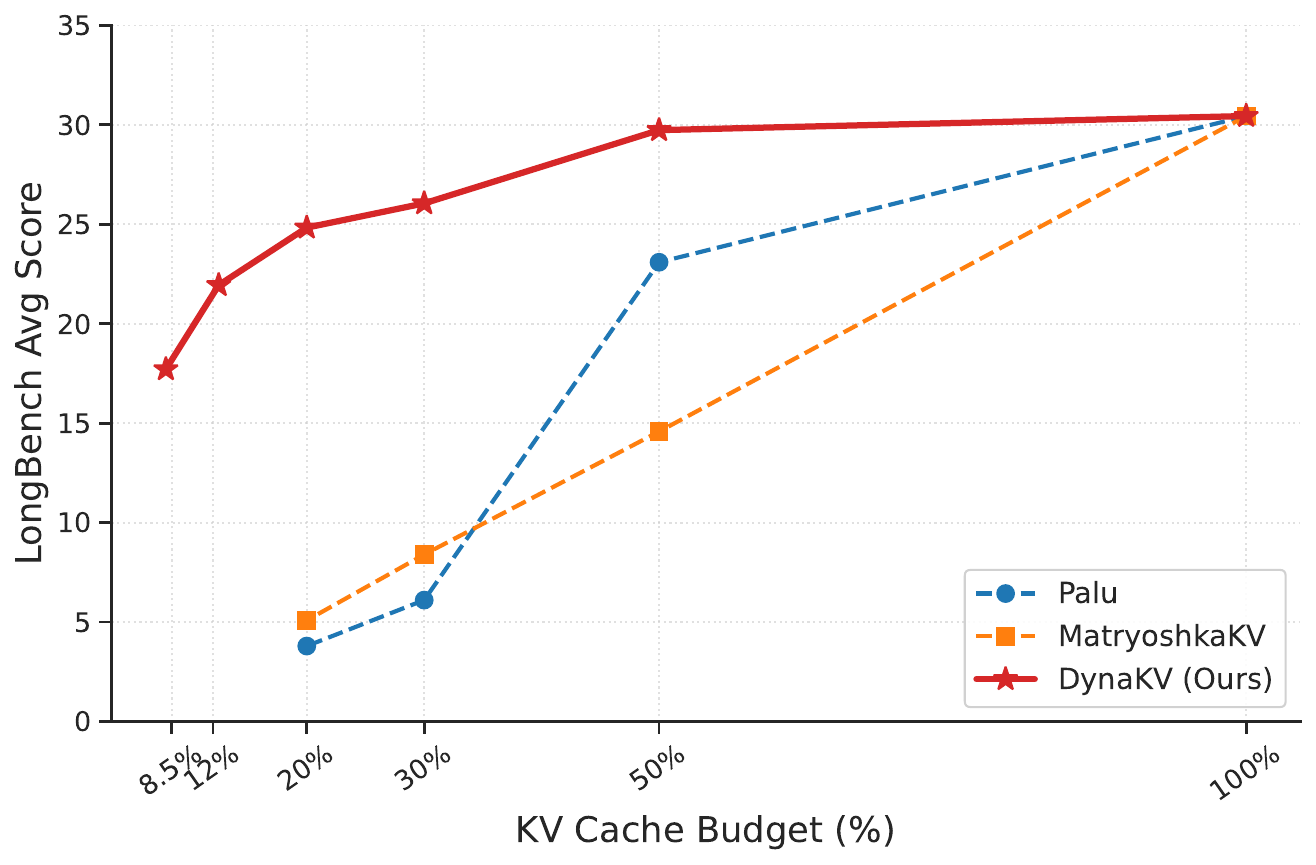}
    
    \vspace{-10pt} % 【可选】收紧图片和 Caption 的距离
    \caption{LongBench average scores under varying KV cache budgets. The plot compares DynaKV with baseline methods (Palu and MatryoshkaKV) across different compression rates.}
    \label{fig:overview}
    \vspace{-10pt} % 【可选】收紧底部空白
\end{wrapfigure}
\paragraph{Performance on Short-Context Tasks}
As shown in Table \ref{tab:short_context}, \textbf{DynaKV} consistently outperforms both Palu and MatryoshkaKV across all settings, and the performance gap widens significantly under aggressive compression regimes. Specifically, at a 20\% retention rate on Llama-3-8B, baseline methods suffer severe degradation, where Palu and MatryoshkaKV drop to 44.99\% and 48.05\%, respectively. In contrast, DynaKV sustains a robust average of 62.08\%, surpassing the strongest baseline by over 14\%. A similar trend is observed on Qwen3-8B; DynaKV dominates Palu (59.17\% vs. 39.66\% at $\sim$18\% rate), validating that our token-adaptive strategy effectively preserves critical semantic information even when the memory budget is strictly limited.

\paragraph{Performance on Long-Context Tasks}

\begin{table*}[t]
\centering
\caption{Detailed performance on LongBench categorized by task type. We report the score for each of the 16 subsets and the overall average (Avg). \textbf{Bold} indicates the best performance under comparable compression rates.}
\label{tab:longbench_categorized}

% 强制缩放以适应页面宽度
\resizebox{\textwidth}{!}{%
    \scriptsize %以此字号为基础进行缩放
    \setlength{\tabcolsep}{2.5pt} % 稍微收紧列间距
    \renewcommand{\arraystretch}{0.8} % 增加行高，避免拥挤

    \begin{tabular}{lc ccc ccc ccc ccc cc cc c}
    \toprule
    % 第一行表头：大类聚合
    \multirow{2}{*}{\textbf{Method}} & \multirow{2}{*}{\textbf{Rate}} & 
    \multicolumn{3}{c}{\textbf{Single-Doc QA}} & 
    \multicolumn{3}{c}{\textbf{Multi-Doc QA}} & 
    \multicolumn{3}{c}{\textbf{Summarization}} & 
    \multicolumn{3}{c}{\textbf{Few-Shot}} & 
    \multicolumn{2}{c}{\textbf{Synthetic}} & 
    \multicolumn{2}{c}{\textbf{Code}} & 
    \multirow{2}{*}{\textbf{Avg}} \\
    
    % 分组横线：(lr) 参数让线左右缩进，视觉更分离
    \cmidrule(lr){3-5} \cmidrule(lr){6-8} \cmidrule(lr){9-11} \cmidrule(lr){12-14} \cmidrule(lr){15-16} \cmidrule(lr){17-18}
    
    % 第二行表头：具体数据集
     & & \textbf{NarQA} & \textbf{Qasp} & \textbf{MF-en} & \textbf{Hotp} & \textbf{2Wiki} & \textbf{Musiq} & \textbf{GovR} & \textbf{QMSum} & \textbf{MNews} & \textbf{TREC} & \textbf{TrivQA} & \textbf{SAM} & \textbf{PasCnt} & \textbf{PasRet} & \textbf{LCC} & \textbf{Repo} & \\
    \midrule
    \multicolumn{19}{c}{\textit{Llama-3-8B}} \\
    \midrule
    Full Cache & 100\% & 5.00 & 18.87 & 20.39 & 9.59 & 11.49 & 6.12 & 27.43 & 23.82 & 2.12 & 75.0 & 90.78 & 44.77 & 2.00 & 9.17 & 72.03 & 68.53 & 30.44 \\
    \midrule
    Palu & 50\% & 7.83 & 16.50 & 17.21 & 9.08 & 9.76 & \textbf{5.27} & 13.49 & 7.00 & 2.31 & \textbf{72.00} & 83.72 & 26.03 & \textbf{1.89} & 3.45 & 47.13 & 46.69 & 23.09 \\
    MatryoshkaKV & 50\% & 1.30 & 13.17 & 11.23 & 3.16 & 7.33 & 2.21 & 12.98 & 7.48 & 15.87 & 48.00 & 19.71 & 10.53 & 1.33 & 3.45 & 38.19 & 37.29 & 14.58 \\
    \textbf{DynaKV} & 47\% & \textbf{19.96} & \textbf{18.84} & \textbf{29.13} & \textbf{9.12} & \textbf{11.62} & 5.22 & \textbf{26.75} & \textbf{22.35} & \textbf{17.95} & 71.50 & \textbf{87.95} & \textbf{41.32} & 0.55 & \textbf{5.53} & \textbf{55.47} & \textbf{52.47} & \textbf{29.73} \\
    \midrule
    Palu & 30\% & 0.70 & 5.69 & 4.57 & 1.33 & 2.02 & 0.67 & 7.11 & 3.14 & 5.39 & 18.50 & 4.79 & 1.24 & 1.29 & \textbf{4.58} & 19.09 & 17.59 & 6.11 \\
    MatryoshkaKV & 30\% & 0.51 & 6.08 & 5.69 & 1.47 & 2.47 & 1.02 & 7.37 & 7.84 & 9.79 & 9.00 & 4.81 & 4.76 & \textbf{1.54} & 3.69 & \textbf{36.49} & 31.84 & 8.40 \\
    \textbf{DynaKV} & 29\% & \textbf{18.18} & \textbf{18.10} & \textbf{29.39} & \textbf{8.97} & \textbf{9.58} & \textbf{4.40} & \textbf{25.73} & \textbf{21.13} & \textbf{10.81} & \textbf{62.50} & \textbf{88.03} & \textbf{40.38} & 1.00 & 3.33 & 33.52 & \textbf{41.70} & \textbf{26.05} \\
    \midrule
    Palu & 20\% & 0.11 & 2.52 & 2.16 & 0.24 & 0.65 & 0.44 & 6.40 & 2.74 & 1.65 & 1.00 & 2.33 & 0.48 & \textbf{2.16} & \textbf{4.33} & 17.75 & 15.81 & 3.80 \\
    MatryoshkaKV & 20\% & 0.71 & 4.41 & 1.92 & 0.68 & 1.28 & 0.95 & 3.18 & 8.51 & 5.04 & 0.00 & 0.92 & 7.21 & 0.94 & 3.24 & 24.07 & 18.35 & 5.09 \\
    \textbf{DynaKV} & 20\% & \textbf{19.80} & \textbf{16.48} & \textbf{24.85} & \textbf{7.95} & \textbf{9.42} & \textbf{3.98} & \textbf{22.96} & \textbf{21.04} & \textbf{10.29} & \textbf{51.00} & \textbf{84.93} & \textbf{37.98} & 0.87 & 4.16 & \textbf{34.80} & \textbf{46.70} & \textbf{24.83} \\
    \midrule
    \textbf{DynaKV} & \definecolor{darkgreen}{rgb}{0.0, 0.5, 0.0} \textcolor{darkgreen}{12\%} & 15.48 & 15.34 & 19.26 & 7.11 & 9.85 & 4.29 & 18.83 & 19.45 & 10.73 & 40.50 & 74.40 & 29.01 & 2.64 & 3.22 & 37.78 & 43.22 & \textbf{21.94} \\
    \textbf{DynaKV} & \definecolor{darkgreen}{rgb}{0.0, 0.5, 0.0} \textcolor{darkgreen}{8.5\%} & 2.63 & 13.46 & 15.82 & 5.18 & 8.53 & 3.74 & 16.22 & 17.38 & 18.41 & 30.50 & 35.54 & 22.64 & 2.95 & 4.43 & 43.67 & 42.31 & \textbf{17.71} \\
    \midrule
    \multicolumn{19}{c}{\textit{Qwen3-8B-Base}} \\
    \midrule
    Full Cache & 100\% & 18.55 & 47.41 & 54.30 & 57.15 & 44.54 & 31.22 & 32.87 & 25.01 & 27.28 & 75.0 & 92.21 & 46.23 & 7.00 & 89.00 & 72.72 & 63.07 & 48.97 \\
    \midrule
    Palu & 50\% & \textbf{13.96} & 16.37 & 28.93 & 9.11 & 10.42 & 5.51 & 15.98 & 8.93 & 10.89 & 65.5 & 83.90 & 19.89 & 3.08 & 3.82 & 20.31 & 25.16 & 21.36 \\
    \textbf{DynaKV} & 48\% & 10.11 & \textbf{23.49} & \textbf{31.08} & \textbf{11.49} & \textbf{11.63} & \textbf{7.05} & \textbf{30.51} & \textbf{24.17} & \textbf{26.76} & \textbf{72.5} & \textbf{91.68} & \textbf{41.79} & \textbf{3.39} & \textbf{56.00} & \textbf{38.41} & \textbf{34.13} & \textbf{32.14} \\
    \midrule
    Palu & 30\% & 1.08 & 6.09 & 9.39 & 3.16 & 5.31 & 2.03 & 11.65 & 3.75 & 8.58 & 41.5 & 12.17 & 3.03 & \textbf{2.96} & 2.95 & 17.21 & 19.99 & 9.43 \\
    \textbf{DynaKV} & 30\% & \textbf{2.51} & \textbf{19.20} & \textbf{21.23} & \textbf{11.44} & \textbf{11.59} & \textbf{8.25} & \textbf{26.29} & \textbf{23.90} & \textbf{25.85} & \textbf{69.5} & \textbf{88.16} & \textbf{38.67} & 2.28 & \textbf{5.67} & \textbf{32.13} & \textbf{34.72} & \textbf{26.34} \\
    \midrule
    Palu & 20\% & 0.79 & 3.73 & 4.20 & 1.61 & 1.70 & 1.39 & 10.72 & 3.53 & 7.15 & 9.5 & 4.19 & 3.26 & \textbf{3.33} & 0.65 & \textbf{17.24} & 15.15 & 5.51 \\
    \textbf{DynaKV} & 18\% & \textbf{2.68} & \textbf{19.20} & \textbf{16.60} & \textbf{9.95} & \textbf{10.21} & \textbf{4.23} & \textbf{20.89} & \textbf{21.21} & \textbf{22.73} & \textbf{60.0} & \textbf{79.98} & \textbf{36.16} & 2.06 & \textbf{5.00} & 32.42 & \textbf{36.40} & \textbf{23.73} \\
    \bottomrule
    \end{tabular}%
}
\end{table*}
\begin{table*}[!ht]
\centering
\caption{Performance on Ruler benchmark. Scores are averaged across context lengths from 4k to 32k. \textbf{Bold} indicates the best performance under comparable compression rates.}
\label{tab:ruler_detailed}

% 使用 resizebox 确保表格完美适配页面宽度
\resizebox{0.8\textwidth}{!}{%
    \scriptsize 
    \setlength{\tabcolsep}{2.5pt} 
    \renewcommand{\arraystretch}{0.8} 

    \begin{tabular}{lc ccc ccccc cc c cc c}
    \toprule
    % 第一行表头：大类聚合
    \multirow{2}{*}{\textbf{Method}} & \multirow{2}{*}{\textbf{Rate}} & 
    \multicolumn{3}{c}{\textbf{Retrieval (Single)}} & 
    \multicolumn{5}{c}{\textbf{Retrieval (Multi)}} & 
    \multicolumn{2}{c}{\textbf{QA}} & 
    \textbf{Reas.} & 
    \multicolumn{2}{c}{\textbf{Agg.}} & 
    \multirow{2}{*}{\textbf{Avg}} \\
    
    % 分组横线
    \cmidrule(lr){3-5} \cmidrule(lr){6-10} \cmidrule(lr){11-12} \cmidrule(lr){13-13} \cmidrule(lr){14-15}
    
    % 第二行表头：具体数据集
     & & \textbf{S-1} & \textbf{S-2} & \textbf{S-3} & \textbf{MK-1} & \textbf{MK-2} & \textbf{MK-3} & \textbf{MQ} & \textbf{MV} & \textbf{Squad} & \textbf{Hotpot} & \textbf{VT} & \textbf{CWE} & \textbf{FWE} & \\
    \midrule
    \multicolumn{16}{c}{\textit{Llama-3-8B}} \\
    \midrule
    Full Cache & 100\% & 50.0 & 50.0 & 50.0 & 49.8 & 49.8 & 49.0 & 49.9 & 49.9 & 37.0 & 28.3 & 49.7 & 49.2 & 44.7 & 46.7 \\
    \midrule
    Palu & 50\% & \textbf{50.0} & 49.5 & 33.3 & 45.8 & \textbf{48.5} & 34.3 & 36.4 & 22.8 & 27.5 & 23.0 & \textbf{45.4} & 28.7 & 32.7 & 36.8 \\
    MatryoshkaKV & 50\% & 49.8 & 46.0 & 24.8 & 16.0 & 1.8 & 0.0 & 7.2 & 3.8 & 7.2 & 11.0 & 0.1 & 6.9 & 3.9 & 13.7 \\
    \textbf{DynaKV} & 47\% & \textbf{50.0} & \textbf{50.0} & \textbf{48.0} & \textbf{46.8} & 47.0 & \textbf{43.3} & \textbf{46.5} & \textbf{46.3} & \textbf{36.3} & \textbf{28.5} & 30.2 & \textbf{36.9} & \textbf{41.5} & \textbf{42.4} \\
    \midrule
    Palu & 30\% & 8.0 & 0.0 & 0.0 & 0.0 & 0.0 & 0.0 & 0.0 & 0.0 & 1.8 & 2.0 & 0.0 & 0.0 & 7.8 & 1.5 \\
    MatryoshkaKV & 30\% & 39.5 & 16.0 & 0.0 & 4.2 & 0.0 & 0.0 & 0.1 & 0.4 & 2.5 & 2.2 & 0.0 & 2.4 & 6.7 & 5.7 \\
    \textbf{DynaKV} & 29\% & \textbf{50.0} & \textbf{50.0} & \textbf{49.3} & \textbf{47.8} & \textbf{47.5} & \textbf{35.0} & \textbf{44.9} & \textbf{41.3} & \textbf{34.3} & \textbf{25.5} & \textbf{29.8} & \textbf{19.8} & \textbf{37.8} & \textbf{39.4} \\
    \midrule
    Palu & 20\% & 0.0 & 0.0 & 0.0 & 0.0 & 0.0 & 0.0 & 0.0 & 0.0 & 0.3 & 0.0 & 0.0 & 0.0 & 0.0 & 0.0 \\
    MatryoshkaKV & 20\% & 0.0 & 0.0 & 0.0 & 0.0 & 0.0 & 0.0 & 0.0 & 0.0 & 0.5 & 0.2 & 0.0 & 0.1 & 2.2 & 0.2 \\
    \textbf{DynaKV} & 20\% & \textbf{49.8} & \textbf{50.0} & \textbf{47.0} & \textbf{46.0} & \textbf{38.8} & \textbf{23.5} & \textbf{39.1} & \textbf{34.9} & \textbf{27.5} & \textbf{24.3} & \textbf{33.6} & \textbf{8.1} & \textbf{39.8} & \textbf{35.6} \\
    \midrule
    \multicolumn{16}{c}{\textit{Qwen3-8B-Base}} \\
    \midrule
    Full Cache & 100\% & 100.0 & 100.0 & 100.0 & 98.8 & 95.0 & 92.0 & 93.9 & 98.0 & 76.3 & 55.3 & 98.2 & 76.6 & 93.8 & 90.6 \\
    \midrule
    Palu & 50\% & 100.0 & 99.3 & 98.0 & 85.3 & 63.0 & 37.0 & 84.5 & 81.6 & 39.0 & 33.3 & 31.4 & 25.6 & 76.0 & 65.7 \\
    \textbf{DynaKV} & 48\% & \textbf{100.0} & \textbf{100.0} & \textbf{98.8} & \textbf{88.3} & \textbf{77.5} & \textbf{66.5} & \textbf{90.1} & \textbf{91.9} & \textbf{70.3} & \textbf{59.5} & \textbf{97.7} & \textbf{45.5} & \textbf{78.5} & \textbf{81.9} \\
    \midrule
    Palu & 30\% & 96.8 & 47.0 & 9.5 & 32.3 & 2.3 & 0.0 & 31.6 & 25.8 & 6.3 & 8.3 & 0.0 & 0.6 & 1.7 & 20.2 \\
    \textbf{DynaKV} & 30\% & \textbf{100.0} & \textbf{99.8} & \textbf{99.0} & \textbf{79.3} & \textbf{75.3} & \textbf{49.8} & \textbf{77.5} & \textbf{78.7} & \textbf{51.3} & \textbf{39.0} & \textbf{92.3} & \textbf{24.4} & \textbf{42.3} & \textbf{69.9} \\
    \midrule
    Palu & 20\% & 9.0 & 0.0 & 0.0 & 1.5 & 0.0 & 0.0 & 1.9 & 2.3 & 0.5 & 1.3 & 0.0 & 0.0 & 0.2 & 1.3 \\
    \textbf{DynaKV} & 18\% & \textbf{100.0} & \textbf{95.8} & \textbf{91.5} & \textbf{57.3} & \textbf{15.8} & \textbf{0.3} & \textbf{62.8} & \textbf{70.6} & \textbf{38.8} & \textbf{35.3} & \textbf{40.8} & \textbf{18.1} & \textbf{40.8} & \textbf{51.3} \\
    \bottomrule
    \end{tabular}%
}
\end{table*}
Figure 3 illustrates the performance trends on LongBench \citep{longbench}. While methods with fixed compression rate degrade rapidly as the memory budget decreases, DynaKV maintains robust understanding even at extreme compression: it achieves 17.71\% with only 8.5\% cache, surpassing Palu's performance at a much larger 30\% budget (6.11\%). This advantage is further amplified on the RULER benchmark \citep{ruler} (Table \ref{tab:ruler_detailed}), where baselines fail completely at a 30\% retention rate (scoring $<$ 6\%), while DynaKV retains a competitive 39.4\%. Similar gains on Qwen3-8B confirm that dynamic capacity allocation effectively prevents context collapse under strict memory constraints.

\paragraph{Perplexity}
Finally, we quantify fundamental information loss via perplexity (PPL) measurements on the C4 and PG-19 datasets (Table \ref{tab:ppl_results}). The results highlight a clear distinction in stability. Baselines like Palu and MatryoshkaKV experience catastrophic PPL surges as the retention rate decreases. For instance, at a 20\% retention rate on Llama-3-8B, Palu's PPL on C4 spikes to 113.90 and Matryoshka to % 确保已引入宏包: \usepackage{wrapfig}

\begin{wraptable}{r}{0.55\columnwidth} % {位置}{给表格预留的总宽度}
    \vspace{-15pt} % 【可选】调整表格与上方文字的垂直距离，负数表示向上提
    \centering
    \caption{Perplexity (PPL) evaluation on C4 and PG-19 datasets. Lower is better. \textbf{Bold} indicates the best performance under comparable compression rates.}
    \label{tab:ppl_results}

    % 这里 resizebox 的宽度改为 \linewidth，意思是填满 wraptable 给定的宽度
    \resizebox{0.7\linewidth}{!}{%
        \scriptsize
        \setlength{\tabcolsep}{3pt} % 稍微调小一点列间距，防止过宽
        \renewcommand{\arraystretch}{0.9} 

        \begin{tabular}{lccc}
        \toprule
        \textbf{Method} & \textbf{Rate} & \textbf{C4} & \textbf{PG-19} \\
        \midrule
        \multicolumn{4}{c}{\textit{Llama-3-8B}} \\
        \midrule
        Full Cache & 100\% & 9.49 & 9.26 \\
        \midrule
        Palu & 50\% & 15.20 & 14.93 \\
        MatryoshkaKV & 50\% & 22.14 & 20.35 \\
        \textbf{DynaKV} & 47\% & \textbf{10.73} & \textbf{10.46} \\
        \midrule
        Palu & 30\% & 48.51 & 53.41 \\
        MatryoshkaKV & 30\% & 43.87 & 38.87 \\
        \textbf{DynaKV} & 29\% & \textbf{11.46} & \textbf{11.34} \\
        \midrule
        Palu & 20\% & 113.90 & 127.88 \\
        MatryoshkaKV & 20\% & 137.59 & 94.64 \\
        \textbf{DynaKV} & 20\% & \textbf{12.51} & \textbf{13.09} \\
        \midrule
        \multicolumn{4}{c}{\textit{Qwen3-8B-Base}} \\
        \midrule
        Full Cache & 100\% & 11.71 & 9.77 \\
        \midrule
        Palu & 50\% & 17.73 & 19.26 \\
        \textbf{DynaKV} & 48\% & \textbf{12.97} & \textbf{11.59} \\
        \midrule
        Palu & 30\% & 47.58 & 50.78 \\
        \textbf{DynaKV} & 30\% & \textbf{14.30} & \textbf{15.30} \\
        \midrule
        Palu & 20\% & 123.51 & 149.38 \\
        \textbf{DynaKV} & 18\% & \textbf{16.01} & \textbf{18.71} \\
        \bottomrule
        \end{tabular}%
    }
    \vspace{-40pt} % 【可选】调整表格与下方文字的距离
\end{wraptable}137.59, indicating severe language degradation. In contrast, DynaKV maintains a low PPL of 12.51 at the same 20\% retention rate. This result verifies that our method effectively preserves core linguistic capabilities even under aggressive compression.

\section{Analysis}
\label{sec:analysis}
\subsection{Retain Rate Allocation}
Figure \ref{fig:bar} and Figure \ref{fig:heatmap} visualizes the token-wise and layer-wise compression rate allocation of DynaKV on a sample sentence: \textit{``I keep telling myself that I am going to start working on it tomorrow, but to be honest, I am just struggling with chronic procrastination."} The bar chart and heatmap reveal three distinct characteristics of our adaptive compression strategy:

\begin{itemize}
    \item \textbf{Dominance of Attention Sinks:} 
    First, we observe a significant retention bias towards the initial token ($<BOS>$). As shown in Figure \ref{fig:heatmap}, it consistently maintains the highest average retention rate ($\sim$0.75) across all layers. This aligns with the ``attention sink" phenomenon observed in previous studies \citep{streamingllm}, where models allocate disproportionate attention mass to initial tokens to stabilize inference. DynaKV successfully identifies and preserves this critical anchor despite its lack of semantic content.
\end{itemize}

%\noindent % 或者留一行空行，或者写一句废话
\vspace{-1em} % 把因为加空行多出来的距离缩回去

\begin{itemize}
    \item \textbf{Semantic-Adaptive Allocation:} 
    Second, the method demonstrates a clear ability to distinguish between high-value semantic tokens and low-information functional ones. 
    Rare or complex words such as \textit{``chronic"} and \textit{``procrastination"} (tokenized as \textit{``procrast"} and \textit{``ination"}) are assigned significantly higher retention rates. 
    In contrast, common functional tokens and stopwords—such as \textit{``that"}, \textit{``to"}, \textit{``be"}, and \textit{``just"}—are aggressively compressed, showing much lower retention rates. This confirms that DynaKV effectively allocates limited memory budget to tokens that contribute most to the semantic density of the context.
\begin{figure*}[t] % [t] 表示强制尝试放在页首
  \centering
  % width=0.95\linewidth 保证图片占满版心宽度但留一点点白边，视觉更舒适
  % 请将 'your_figure_filename.pdf' 替换为您实际的文件名
  \includegraphics[width=1\linewidth]{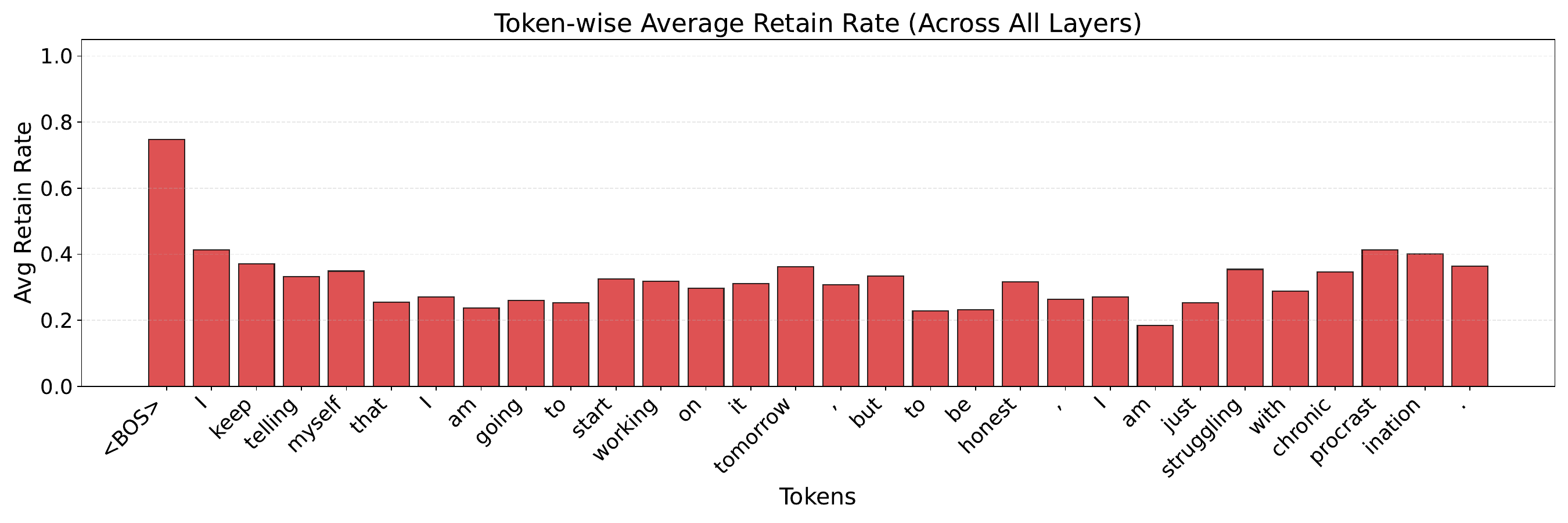}
  
  \vspace{-1em} % 如果图片和标题距离太远，可以用这个负值拉近一点
  
\caption{Token-wise average retention rate across all layers. The bar height represents the budget allocated to each token, highlighting the high retention of the sink token and semantic tokens compared to functional ones.}
  \label{fig:bar}
  
  \vspace{-1em} % 减少标题下方的空白，节省版面
\end{figure*}
\begin{figure*}[t] % [t] 表示强制尝试放在页首
  \centering
  % width=0.95\linewidth 保证图片占满版心宽度但留一点点白边，视觉更舒适
  % 请将 'your_figure_filename.pdf' 替换为您实际的文件名
  \includegraphics[width=0.95\linewidth]{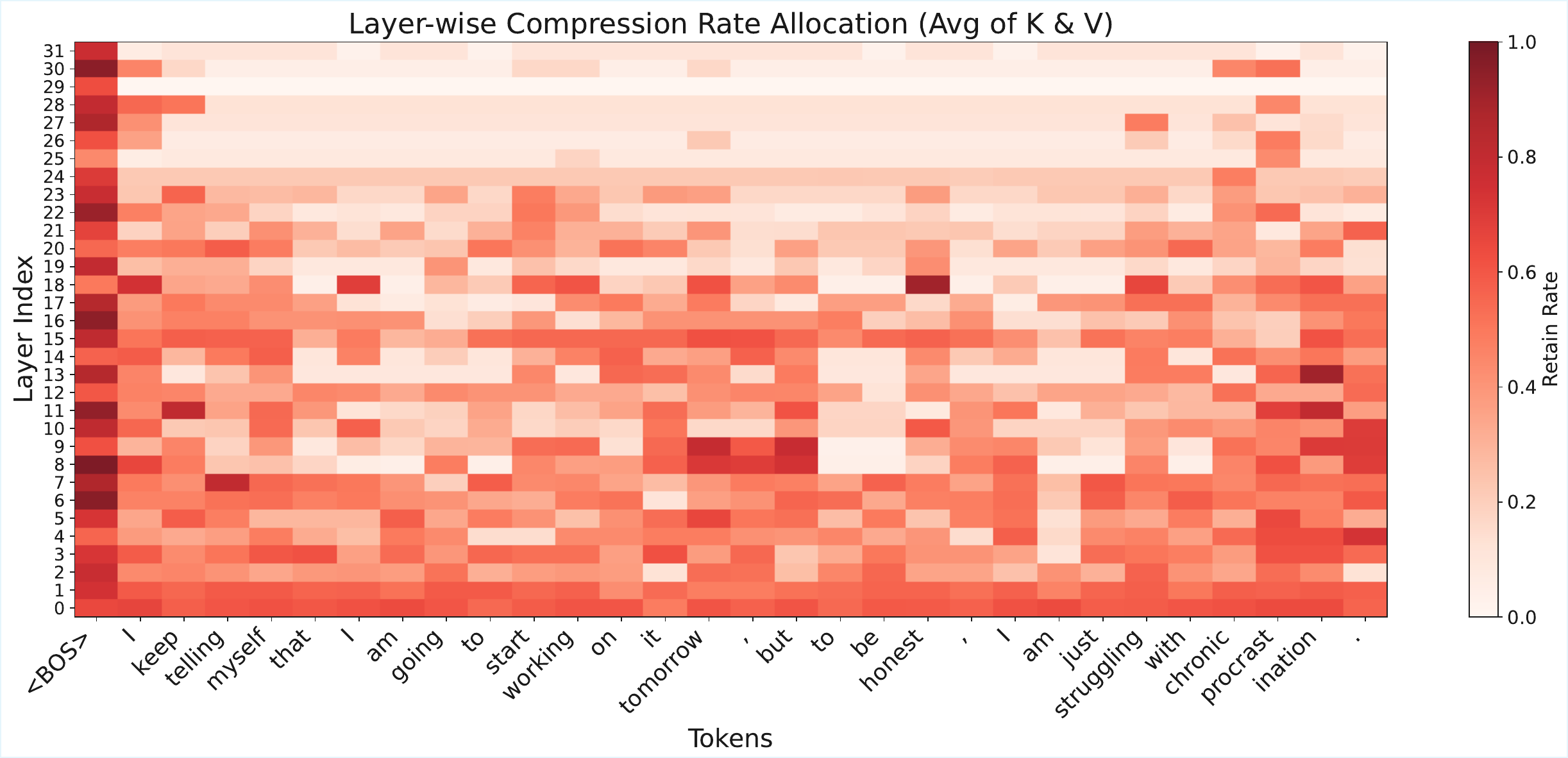}
  
  \vspace{-1em} % 如果图片和标题距离太远，可以用这个负值拉近一点
  
  \caption{Layer-wise compression rate allocation heatmap. The x-axis denotes input tokens and the y-axis denotes model layers (0--31). Darker red indicates a higher retention rate, visualizing the dynamic distribution of memory budgets across different network depths.}
  \label{fig:heatmap}
  
  \vspace{-1em} % 减少标题下方的空白，节省版面
\end{figure*}

    \item \textbf{Hierarchical Layer-wise Distribution:} 
    Third, Figure \ref{fig:heatmap} reveals a distinct hierarchical pattern in compression intensity. The lower layers exhibit generally higher retention rates, suggesting that early layers prioritize preserving a broader range of syntactic information. Conversely, the deeper layers show a shift towards lighter colors, indicating higher compression rates. This suggests that as the model abstracts information in higher layers, it requires fewer tokens to maintain context, allowing DynaKV to compress more aggressively with low performance loss.
\end{itemize}

\subsection{Integration with Sequence-Dimension Compression}
Our DynaKV compresses the Key-Value cache along the channel dimension, making it orthogonal to sequence-level pruning methods like SnapKV \citep{snapkv}. To evaluate their compatibility, we integrated DynaKV with SnapKV, with detailed results in Table \ref{tab:snapkv}. The combined approach proves remarkably robust, retaining 94\% performance even at a strict 6\% budget. This confirms that our spectral compression synergizes effectively with token sparsity mechanisms, enabling extreme memory efficiency without compromising model capability.

\paragraph{Implementation Details}
It is important to note a specific constraint in this integration. Since DynaKV performs a joint spectral decomposition across all attention heads to maximize global information retention, the compressed latent representations are shared. Consequently, we cannot perform head-specific token eviction. Instead, we adopt a \textit{global token eviction strategy}: we aggregate importance scores across all heads and apply a unified mask to evict identical tokens for every head. Despite this coarser granularity compared to per-head pruning, the experimental results confirm that the semantic density provided by DynaKV compensates for the lack of head-specific flexibility.
\begin{table}[!h]
\centering
\caption{Synergistic performance of DynaKV integrated with SnapKV on LongBench (Llama-3-8B). \textbf{Snap($L$)+DynaKV~$R$\%} denotes applying SnapKV to retain $L$ tokens, followed by DynaKV compression at rate $R$\%. This hybrid approach achieves extreme compression (down to $\sim$3.7\%) while maintaining robust performance.}
\label{tab:snapkv}

\resizebox{0.6\columnwidth}{!}{%
    \setlength{\tabcolsep}{8pt}
    \renewcommand{\arraystretch}{1.25}
    
    \begin{tabular}{l c c c}
    \toprule
    \textbf{Method Configuration} & \textbf{Base Budget} & \textbf{Retain Rate} & \textbf{LongBench} \\
    \midrule
    Full Cache & 8192 & 100.0\% & 30.44 \\
    \midrule
    SnapKV (2048) & 2048 & 25.0\% & 29.95 \\  % 新增行
    Snap(2048) + DynaKV 47\% & 2048 & $\sim$11.8\% & 29.19 \\
    Snap(2048) + DynaKV 30\% & 2048 & $\sim$7.5\% & 25.60 \\
    \midrule
    SnapKV (1024) & 1024 & 12.5\% & 29.56 \\  % 新增行
    Snap(1024) + DynaKV 47\% & 1024 & $\sim$5.9\% & 28.88 \\
    Snap(1024) + DynaKV 30\% & 1024 & $\sim$3.7\% & 25.08 \\
    \bottomrule
    \end{tabular}%
}
\end{table}

\subsection{Decoding Latency}
While DynaKV significantly reduces the memory footprint, it introduces a marginal computational overhead, primarily stemming from the spectral gating mask calculation and the reconstruction of Key-Value pairs during attention computation. Our method maintains approximately 85\% of the end-to-end throughput compared to the Full Cache baseline. This 15\% latency cost is a strategic trade-off for overcoming the physical memory limitations of hardware. In long-context generation, the bottleneck is predominantly memory capacity rather than compute speed; thus, the slight reduction in throughput is justified by the capability to process extensive contexts on memory-constrained devices. Consequently, DynaKV provides a practical solution for memory-bound scenarios, enabling high-quality reasoning and long-context generation.

\section{Conclusion}
In this work, we introduce DynaKV, a novel post-training framework for low-rank KV cache compression. To the best of our knowledge, DynaKV is the first approach to move beyond the rigid ``one-size-fits-all" compression paradigm by implementing a \textbf{token-adaptive} strategy. By dynamically allocating compression rates based on the semantic significance of individual tokens, our method effectively resolves the dilemma between compression ratio and model performance that limits existing techniques.

Extensive experiments validate that DynaKV can be seamlessly adapted to pre-trained LLMs with minimal training overhead, achieving significant memory reduction while maintaining robust generation quality where uniform compression methods fail. Furthermore, we demonstrate that our dimensionality reduction approach is orthogonal to sequence-level pruning, allowing for synergistic integration with methods like SnapKV to push the boundaries of memory efficiency. We hope this work inspires further exploration into adaptive, semantics-aware compression strategies for scalable LLM deployment.

% ====================
% BIBLIOGRAPHY
% ====================
\bibliography{main}

\end{document}